\documentclass[11pt]{article}

\usepackage[final]{acl}

\usepackage{url}
\usepackage[table]{xcolor}
\definecolor{directbg}{RGB}{232,242,255}
\definecolor{fixedbg}{RGB}{235,248,238}
\definecolor{optbg}{RGB}{255,242,224}

\usepackage[most]{tcolorbox}
\usepackage{listings}
\usepackage{xcolor}

\definecolor{promptbg}{RGB}{248,248,248}
\definecolor{promptframe}{RGB}{180,180,180}
\usepackage{times}
\usepackage{latexsym}
\usepackage[T1]{fontenc}
\usepackage[utf8]{inputenc}
\usepackage{microtype}
\usepackage{inconsolata}
\usepackage{graphicx}
\usepackage{booktabs}
\usepackage[table]{xcolor}
\usepackage{multirow}
\usepackage{amsmath}
\usepackage{amssymb}
\usepackage{enumitem}
\usepackage{tabularx}
\usepackage{array}
\usepackage{adjustbox}

\newcommand{\basewin}{\rlap{\ensuremath{^\dagger}}}

\title{MemPro: Agentic Memory Systems as Evolvable Programs}

\author{
{\bfseries
Qingshan Liu$^{1,*}$ \quad
Guoqing Wang$^{1,*}$ \quad
Wen Wu$^{1,\dagger}$ \quad
Jingqi Huang$^{1}$
} \\
{\bfseries
Xinqi Tao$^{2}$ \quad
Dejia Song$^{2}$ \quad
Jie Zhou$^{1}$ \quad
Liang He$^{1}$
} \\
$^{1}$East China Normal University \\
$^{2}$Xiaohongshu Inc. \\
\texttt{\{51285901015,wgq\}@stu.ecnu.edu.cn} \quad
\texttt{wwu@cs.ecnu.edu.cn}
}

\begin{document}
\maketitle

\renewcommand{\thefootnote}{\fnsymbol{footnote}}
\footnotetext[1]{Equal contribution.}
\footnotetext[2]{Corresponding author.}
\renewcommand{\thefootnote}{\arabic{footnote}}

\begin{abstract}
Long-horizon autonomous agents require memory systems to retain historical information, track evolving states, and reuse relevant knowledge beyond finite context windows. Existing agentic memory systems typically follow a memory construction--retrieval (MCR) pipeline, but often adapt mainly the memory bank while keeping the surrounding pipeline fixed after deployment. This fixed-pipeline design struggles to handle heterogeneous task-specific failure modes and can become misaligned with memory banks that evolve in scale and structure over time. To address these limitations, we propose \textbf{MemPro}, a system-level evolution framework that treats the entire MCR pipeline as an evolvable program rather than adapting only the memory bank or prompt text. MemPro maintains a version tree of runnable memory-system implementations, where an \textit{Evolving Agent} iteratively selects promising versions, diagnoses recurring failures, and creates improved child versions through failure-mode-guided edit--debug refinement. Experiments on LongMemEval, LoCoMo, HotpotQA, and NarrativeQA show that MemPro consistently outperforms strong static and prompt-level evolving baselines within a few iterations, continues to improve with evolution, and achieves a favorable performance--cost trade-off. Code is available at \url{https://github.com/wanghai673/MemPro}.
\end{abstract}

\section{Introduction}

\begin{figure*}[t!]
\centering
\includegraphics[width=0.95\textwidth]{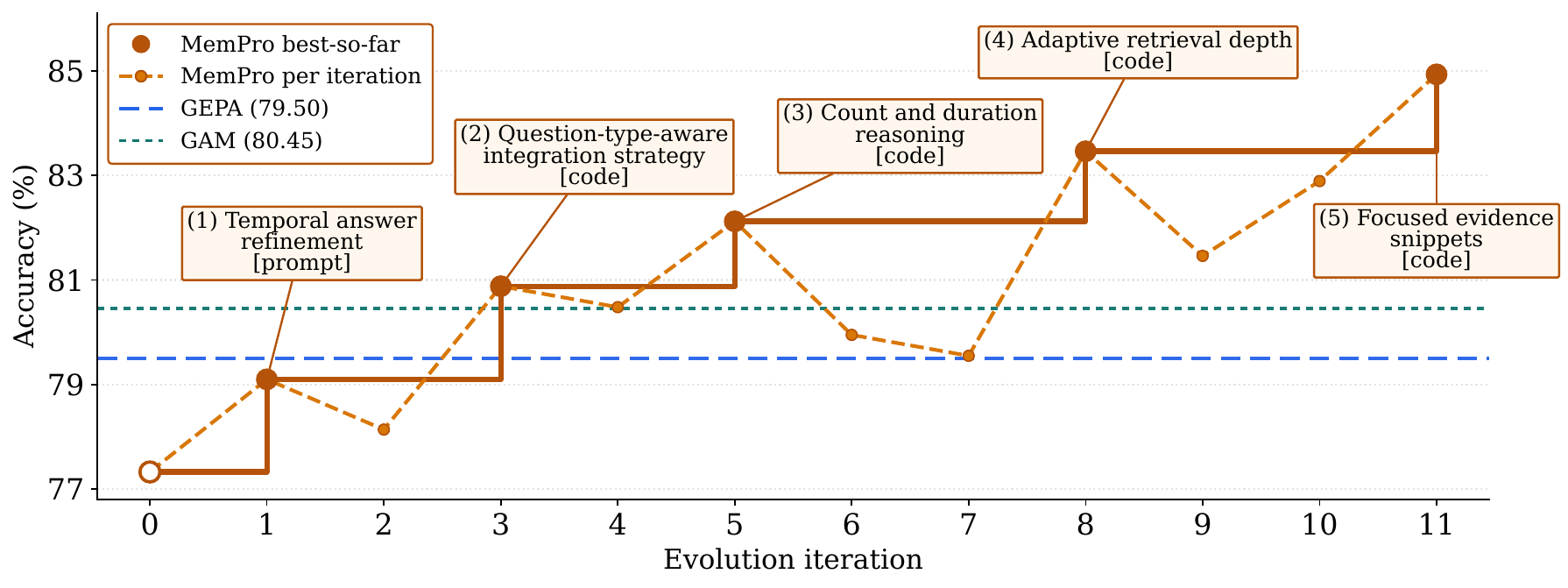}
\caption{
Evolution dynamics of MemPro on the LoCoMo evaluation set, showing the performance of each evolved version, the best-so-far performance, and the main improvements of performance-enhancing versions.
}

\label{fig:evolution_dynamics}
\vspace{-8pt}
\end{figure*}

Large language models (LLMs)~\citep{gpt3, gpt4, llama3, qwen3} increasingly serve as the foundation for autonomous agents~\citep{wang2024survey}, yet long-horizon tasks and sustained interactions require them to retain and reuse historical information over time. Simply extending the context window with more history is costly, noisy, and insufficient for maintaining structured long-term state~\citep{memgpt}. Memory systems therefore play a central role, maintaining and retrieving task- or user-relevant information beyond the context window~\citep{zhang2025survey, amem}.

Recent agentic memory systems typically follow a memory construction--retrieval (MCR) pipeline, where construction builds or updates a structured memory bank from interaction histories or task inputs, and retrieval selects and uses relevant memories to answer downstream queries~\citep{gam, mem0}. Prior work has improved this pipeline by designing more structured and compact memory banks, including hierarchical organizations~\citep{memgpt, memoryos}, graph-based or dynamically linked memories~\citep{amem, mem0}, and summarization or compression pipelines~\citep{memorybank, lightmem}. However, existing agentic memory systems typically treat the memory bank as the primary adaptive component, while the surrounding MCR pipeline is manually designed and kept fixed after deployment~\citep{zhang2025survey}. This fixed-pipeline assumption leads to two key limitations. First, it struggles with \textit{task heterogeneity}: different long-term memory tasks exhibit different failure modes and therefore require different memory-use strategies. For example, temporal reasoning may require tracking event order or identifying the latest state, whereas multi-session reasoning may require linking scattered evidence across sessions—yet a fixed pipeline must apply the same strategy to all of them. Second, it creates \textit{memory--pipeline misalignment}: as the memory bank evolves in scale and structure over time, a fixed pipeline may no longer match how memories are organized and used. Together, these limitations can lead to incomplete retrieval, noisy evidence, or ineffective use of retrieved memories. Although prompt-level optimization methods~\citep{gepa, dspy} can adapt the textual components of such systems, they cannot change the executable pipeline logic and are therefore insufficient to address the above limitations. This motivates a broader view of memory self-evolution: an agentic memory system should not only update the memory bank or prompt text, but also evolve the MCR pipeline system-wide.

To address these limitations, we propose \textbf{MemPro} (Agentic \textbf{Mem}ory Systems as Evolvable \textbf{Pro}grams), a system-level evolution framework that treats the MCR pipeline as an evolvable program. MemPro evolves runnable memory-system versions containing both prompts and executable code for constructing and maintaining memory banks, as well as using retrieved memories to solve downstream queries. It maintains a version tree of MCR pipeline implementations, where each node corresponds to a runnable pipeline version and its evaluation log. Starting from an initial pipeline, an \textit{Evolving Agent} iteratively selects promising versions, diagnoses recurring failure modes, and creates improved child versions through failure-mode-guided edit--debug refinement. This tree structure lets MemPro branch from strong historical versions and explore alternative directions rather than following a single linear trajectory. \autoref{fig:evolution_dynamics} illustrates the evolution dynamics of MemPro.

We evaluate MemPro on two long-term memory benchmarks, LongMemEval and LoCoMo, and two long-context QA benchmarks, HotpotQA and NarrativeQA. Across both memory-centric and QA settings, MemPro consistently outperforms strong static and prompt-level evolving baselines within a few iterations, and keeps improving as the version tree expands—suggesting that evolving the executable MCR pipeline yields benefits beyond adapting the memory bank or prompts alone. Our contributions are as follows:
\begin{itemize}
    \item We identify two limitations of fixed-pipeline agentic memory systems: task heterogeneity and memory--pipeline misalignment. We argue that memory self-evolution should operate at the system level rather than only on stored memories or prompt text.
    \item We propose \textbf{MemPro}, a system-level evolution framework that treats the MCR pipeline as an evolvable program. MemPro maintains a version tree of runnable pipeline implementations and uses failure-mode-guided edit--debug refinement to evolve both prompts and executable pipeline code.
    \item We conduct extensive experiments across four long-term memory and long-context QA benchmarks, demonstrating MemPro's consistent gains over strong baselines, continued improvement with evolution, and favorable performance--cost trade-off.
\end{itemize}

\section{Related Work}

\subsection{Agentic Memory Systems}

Agentic memory systems extend LLM agents beyond finite context windows and typically follow a memory construction--retrieval pipeline, where a memory bank is built or updated from historical inputs and later retrieved for downstream tasks. Prior work has improved this pipeline through persistent memory management~\citep{park2023generative, memorybank, memgpt}, structured or hierarchical memory organization~\citep{memoryos, timem}, graph-based or dynamically linked memories~\citep{mem0, amem}, lightweight summarization and compression~\citep{lightmem, simplemem, gam}, learned or heuristic memory writing and retrieval strategies~\citep{memoryr1, memalpha, agemem}, and experience-based or procedural memory reuse~\citep{awm, memp, reme, reasoningbank}. 

Despite these advances, most systems adapt mainly the memory bank while keeping the surrounding pipeline fixed after deployment, so it struggles with heterogeneous, task-specific failure modes and may become misaligned with evolving memory banks. In contrast, MemPro treats the whole pipeline as an evolvable program and optimizes runnable memory-system implementations.

\subsection{Prompt-Level Evolution}

Prompt-level evolution methods improve LLM systems by refining textual instructions without updating model weights. Representative methods optimize prompts through instruction search, feedback, or evolutionary refinement~\citep{ape, opro, protegi, textgrad, promptbreeder, dspy}. GEPA further uses trajectory-level reflection to diagnose failures and evolve prompts~\citep{gepa}. Closely related to agentic memory, MetaMem optimizes a self-evolving meta-memory that provides textual guidance for using memorized knowledge~\citep{metamem}.

When applied to memory systems, prompt-level evolution can refine prompts and improve over static systems, but cannot modify the executable logic that constructs memory banks or uses retrieved memories to answer queries. MemPro instead evolves runnable pipeline implementations—both prompts and executable code—enabling system-level self-evolution.

\begin{figure*}[t!]
\centering
\includegraphics[width=\textwidth]{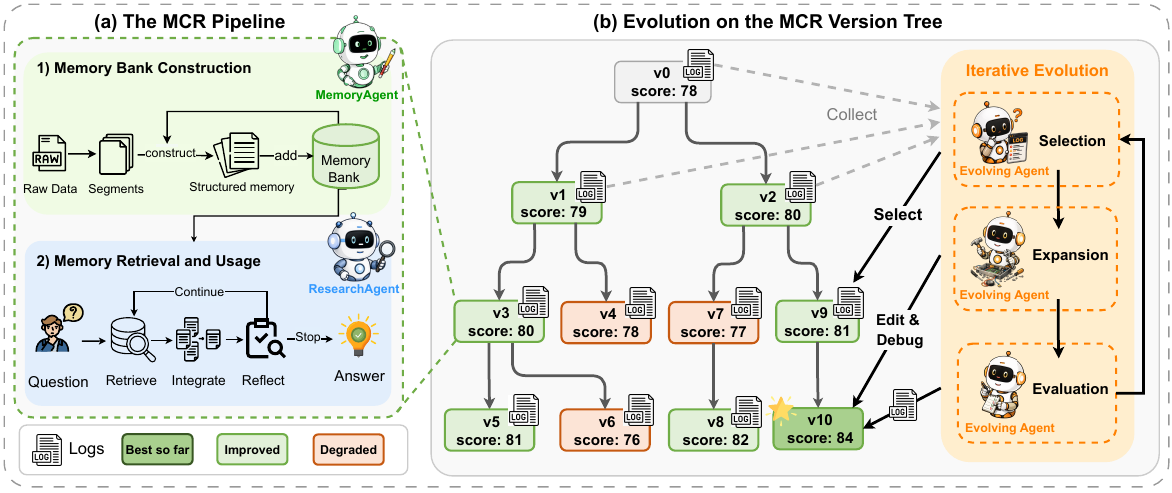}
\caption{
Overview of MemPro. (a) The MCR pipeline. (b) MemPro performs evolution over a version tree: the \textit{Evolving Agent} selects a node based on logs, expands it into a new version, and generates its evaluation log.
}
\label{fig:main}
\vspace{-8pt}
\end{figure*}

\section{Preliminaries}
We introduce the Memory Construction--Retrieval (MCR) pipeline, which captures a common structure in recent agentic memory systems. The MCR pipeline consists of two stages: (1) memory bank construction and (2) memory retrieval and usage. \autoref{fig:main} (left) illustrates the MCR pipeline.
\paragraph{Memory Bank Construction.} At time step \(t\), the raw data \(D_t\) are first segmented into structured segments
\(\mathcal{S}_t=\{s_i^t\}_{i=1}^{N_t}\), where \(s_i^t\) denotes the \(i\)-th segment, such as a dialogue session or document chunk, and \(N_t\) is the number of segments. Given \(\mathcal{S}_t\), the previous memory bank \(\mathcal{M}_{t-1}\), and the memory construction prompt \(I_{\mathrm{mem}}\) in~\autoref{fig:memoryagent_prompt}, a \textit{Memory Agent} \(\mathcal{A}_{\mathrm{mem}}\) produces memory updates
\(\Delta \mathcal{M}_t = \mathcal{A}_{\mathrm{mem}}(I_{\mathrm{mem}}, \mathcal{S}_t, \mathcal{M}_{t-1})\), which are incorporated into the memory bank to obtain
\(
\mathcal{M}_t = \textsc{Update}(\mathcal{M}_{t-1}, \Delta \mathcal{M}_t).
\)

\paragraph{Memory Retrieval and Usage.} Given a query \(q\) and the memory bank \(\mathcal{M}\), a \textit{Research Agent} \(\mathcal{A}_{\mathrm{res}}\) iteratively retrieves and uses memory information to solve the query. It maintains a research state \(z_k\), which denotes the accumulated key information relevant to answering \(q\), with \(z_0=\emptyset\). For each iteration \(k=1,\ldots,K\), the agent performs three steps under different system prompts. \textbf{(1) Retrieval.} Under the retrieval prompt \(I_{\mathrm{ret}}\) in~\autoref{fig:retrieval_prompt}, the agent examines \(q\) and the current research state \(z_{k-1}\) to determine what information is still needed, and generates a retrieval request \(r_k = \mathcal{A}_{\mathrm{res}}(I_{\mathrm{ret}}, q, z_{k-1})\). The memory bank then returns relevant information \(u_k = \textsc{Retrieve}(r_k, \mathcal{M})\), where \(u_k\) denotes the returned memory information. \textbf{(2) Integration.} Under the integration prompt \(I_{\mathrm{int}}\) in~\autoref{fig:integration_prompt}, the agent integrates the returned information with the query and the previous research state to update the research state, \(z_k = \mathcal{A}_{\mathrm{res}}(I_{\mathrm{int}}, q, z_{k-1}, u_k)\). \textbf{(3) Reflection.} Under the reflection prompt \(I_{\mathrm{ref}}\) in~\autoref{fig:reflection_prompt}, the agent judges whether the current research state \(z_k\) contains sufficient information to answer \(q\), denoted by \(b_k = \mathcal{A}_{\mathrm{res}}(I_{\mathrm{ref}}, q, z_k)\), where \(b_k \in \{\textsc{Continue}, \textsc{Stop}\}\). If \(b_k=\textsc{Stop}\) or the maximum number of retrieval steps \(K\) is reached, the agent generates the final answer \(\hat{y} = \mathcal{A}_{\mathrm{res}}(I_{\mathrm{ans}}, q, z_k)\), where \(I_{\mathrm{ans}}\) denotes the answer-extraction prompt; otherwise, it continues.

\section{Methodology}
Our objective is to enable system-level evolution by treating the entire MCR pipeline as the optimization target. To this end, we propose MemPro, which goes beyond updating the memory bank or modifying prompts and instead optimizes the MCR pipeline as an evolvable program through iterative failure-mode-guided refinement. \autoref{fig:main} gives an overview of MemPro.

\subsection{MCR Version Tree}

\paragraph{Overview.}
In MemPro, we construct and maintain an MCR version tree
\(\mathcal{T}=(\mathcal{V},\mathcal{E})\), where each node
\(v\in\mathcal{V}\) represents a runnable implementation of an MCR pipeline
\(F_v\) together with its evaluation log \(L_v\). The evaluation log serves as the basis for subsequent evolution. Each non-root node \(v\) is derived from its parent \(\mathrm{pa}(v)\), while the root \(v_0\) stores the initial MCR pipeline \(F_{v_0}\) and log \(L_{v_0}\).

\paragraph{Evaluation Log.}
We first split a small training set $\mathcal{D}_{\mathrm{train}}$ from the evaluation dataset and use it to guide evolution; the detailed splitting strategy is described in~\ref{sec:exp_setup}. For an MCR pipeline version \(F_v\), the evaluation log \(L_v=(S_v,C_v,\mathrm{pa}(v),A_v)\) records its performance and diagnostic information on the training set. Detailed execution traces are stored in the version registry and used for diagnostic analysis. Here, \(S_v\) denotes the overall score, i.e., the average score on the training set, and serves as the primary metric for assessing the strength of version \(v\). \(C_v\) denotes category-level scores, measuring the average score of \(v\) on each fine-grained category and indicating where the version is weak. \(\mathrm{pa}(v)\) denotes the parent node from which \(v\) evolves, with the root node having no parent. \(A_v\) denotes the overall assessment, which summarizes the major failure modes of \(v\) and possible improvement directions. The root log \(L_{v_0}\) is obtained by evaluating the initial pipeline \(F_{v_0}\) on \(\mathcal{D}_{\mathrm{train}}\).

\subsection{Evolution on the MCR Version Tree}

Given the number of evolution iterations \(T\), MemPro starts from the root node \(v_0\) and performs iterative evolution on the MCR version tree \(\mathcal{T}=(\mathcal{V},\mathcal{E})\) for \(\ell=1,\ldots,T\). The evolution process is controlled by an \textit{Evolving Agent} \(\mathcal{A}_{\mathrm{evo}}\). Each iteration consists of three steps: (1) selection, which chooses the parent node for expansion, (2) expansion, which creates a new version from the selected node, and (3) evaluation, which evaluates the new pipeline version. We detail these steps below.

\paragraph{Selection.}
The \textit{Evolving Agent} collects all evaluation logs in the version tree and compares them to select the most promising node for expansion. In addition to the selected node, it also produces a diagnostic analysis that summarizes the logs in the current tree and analyzes the failure modes and improvement directions of the selected node. Formally, given the selection prompt \(I_{\mathrm{sel}}\) and all logs \(\{L_v\}_{v\in\mathcal{V}}\), the agent outputs
\begin{equation}
    (v_\ell^\star, R_\ell)
    =
    \mathcal{A}_{\mathrm{evo}}(I_{\mathrm{sel}}, \{L_v\}_{v\in\mathcal{V}}),
\end{equation}
where \(v_\ell^\star\) denotes the selected node at evolution iteration \(\ell\), and \(R_\ell\) denotes the diagnostic analysis. The selection is guided by three criteria: \textit{high overall score}, \textit{generalizable improvement directions}, and \textit{frequent failure modes}. More details are specified in the selection prompt \(I_{\mathrm{sel}}\) in~\autoref{app:selection_prompt}.

\paragraph{Expansion.}
Given the selected node \(v_\ell^\star\), its MCR pipeline version \(F_{v_\ell^\star}\), the diagnostic analysis \(R_\ell\), and the expansion prompt \(I_{\mathrm{exp}}\) shown in~\autoref{app:exp_prompt}, the \textit{Evolving Agent} initializes the new version from the selected version, i.e., \(F_{u_\ell}^{(0)}=F_{v_\ell^\star}\) and \(R_\ell^{(0)}=R_\ell\). Here, \(R_\ell^{(j)}\) denotes the accumulated diagnostic analysis at inner expansion step \(j\). The expansion stage proceeds iteratively. At step \(j\), the agent chooses one action \(a_j\in\{\textsc{Edit}, \textsc{Debug}, \textsc{Terminate}\}\):
\begin{equation}
a_j
= \mathcal{A}_{\mathrm{evo}}\bigl(
I_{\mathrm{exp}},
F_{u_\ell}^{(j)},
R_\ell^{(j)}
\bigr).
\end{equation}
If \(a_j=\textsc{Edit}\), the agent revises the current MCR pipeline according to the accumulated diagnostic analysis \(R_\ell^{(j)}\), while keeping \(R_\ell^{(j+1)}=R_\ell^{(j)}\):
\begin{equation}
F_{u_\ell}^{(j+1)}
= \mathcal{A}_{\mathrm{evo}}\!\left(
I_{\mathrm{exp}}, F_{u_\ell}^{(j)}, R_\ell^{(j)}
\right).
\end{equation}
If \(a_j=\textsc{Debug}\), the agent selects a diagnostic example based on the failure analysis in \(R_\ell^{(j)}\), avoiding full training-set re-evaluation:
\begin{equation}
x_j^{\mathrm{diag}}
=
\mathcal{A}_{\mathrm{evo}}\bigl(
I_{\mathrm{exp}},
R_\ell^{(j)},
\mathcal{D}_{\mathrm{train}}
\bigr).
\end{equation}
The current version \(F_{u_\ell}^{(j)}\) is executed on this example to obtain a full trace \(\tau_j=F_{u_\ell}^{(j)}(x_j^{\mathrm{diag}})\). 
The trace updates the diagnostic analysis while keeping the pipeline unchanged, \(F_{u_\ell}^{(j+1)}=F_{u_\ell}^{(j)}\):
\begin{equation}
R_\ell^{(j+1)}
= \mathcal{A}_{\mathrm{evo}}\!\left(
I_{\mathrm{exp}},
R_\ell^{(j)},
x_j^{\mathrm{diag}},
\tau_j
\right).
\end{equation}
This targeted debugging avoids re-evaluating the full \(\mathcal{D}_{\mathrm{train}}\) during each inner expansion step, improving efficiency.
If \(a_j=\textsc{Terminate}\), the expansion process stops and the current version is taken as the new pipeline version, \(F_{u_\ell}=F_{u_\ell}^{(j)}\).
The process repeats until the agent outputs \(\textsc{Terminate}\) or the maximum number of expansion steps \(J\) is reached; in the latter case, \(F_{u_\ell}=F_{u_\ell}^{(J)}\).

\paragraph{Evaluation.}
After obtaining the new pipeline version \(F_{u_\ell}\), we execute it on the full training set \(\mathcal{D}_{\mathrm{train}}\) to collect execution traces:
\begin{equation}
\Gamma_{u_\ell}
=
\{\, F_{u_\ell}(x) \mid x \in \mathcal{D}_{\mathrm{train}} \,\}.
\end{equation}
The \textit{Evolving Agent} then summarizes these traces into an evaluation log:
\begin{equation}
L_{u_\ell}
=
\mathcal{A}_{\mathrm{evo}}
\!\left(
I_{\mathrm{eval}},
F_{u_\ell},
\Gamma_{u_\ell}
\right),
\end{equation}
where \(I_{\mathrm{eval}}\) denotes the evaluation prompt, which is provided in~\autoref{app:eval_prompt}. The resulting log \(L_{u_\ell}=(S_{u_\ell}, C_{u_\ell}, \mathrm{pa}(u_\ell), A_{u_\ell})\) records the training performance, category-level performance, parent node, and overall assessment of the new pipeline version. The pair \((F_{u_\ell}, L_{u_\ell})\) defines a new node \(u_\ell\) with \(\mathrm{pa}(u_\ell)=v_\ell^\star\), which is added to \(\mathcal{T}\).

\definecolor{directbg}{RGB}{232,242,255}
\definecolor{fixedbg}{RGB}{235,248,238}
\definecolor{optbg}{RGB}{255,242,224}

\definecolor{archbg}{RGB}{232,242,255}
\definecolor{promptbg}{RGB}{236,248,233}
\definecolor{MemProbg}{RGB}{255,242,218}

\definecolor{avgbg}{RGB}{255,248,220}

\begin{table*}[t]
\centering
\small

\resizebox{\textwidth}{!}{
\begin{tabular}{lcccccccccccc}
\toprule
\multirow{2}{*}{\textbf{Method}} 
& \multicolumn{7}{c}{\textbf{LongMemEval}} 
& \multicolumn{5}{c}{\textbf{LoCoMo}} \\
\cmidrule(lr){2-8} \cmidrule(lr){9-13}
& \textbf{Temp} & \textbf{Multi} & \textbf{Know} & \textbf{User} & \textbf{Asst.} & \textbf{Pref.} & \cellcolor{avgbg}\textbf{Avg.}
& \textbf{Multi} & \textbf{Open} & \textbf{Single} & \textbf{Temp} & \cellcolor{avgbg}\textbf{Avg.} \\
\midrule

\rowcolor{gray!12}
\multicolumn{13}{l}{\textbf{\textit{GPT-4o-mini}}} \\
\midrule

Full Text 
& 31.58 & 45.45 & 76.92 & 87.14 & 89.29 & 36.67 & \cellcolor{avgbg}56.89
& 68.79 & 56.25 & 86.56 & 50.16 & \cellcolor{avgbg}73.83 \\

RAG 
& 39.85 & 48.48 & 67.95 & 90.00 & \textbf{98.21} & 53.33 & \cellcolor{avgbg}60.90
& 55.32 & 47.92 & 70.99 & 56.39 & \cellcolor{avgbg}63.64 \\

\cellcolor{archbg}LangMem 
& 15.79 & 20.30 & 66.67 & 60.00 & 46.43 & 60.00 & \cellcolor{avgbg}37.20
& 52.10 & 41.65 & 62.80 & 43.25 & \cellcolor{avgbg}55.45 \\

\cellcolor{archbg}Mem0 
& 40.15 & 46.21 & 70.12 & 81.43 & 41.07 & 60.00 & \cellcolor{avgbg}53.51
& 30.85 & 34.38 & 38.41 & 37.07 & \cellcolor{avgbg}36.49 \\

\cellcolor{archbg}A-MEM 
& 47.36 & 48.87 & 64.11 & 92.86 & \underline{96.43} & 46.67 & \cellcolor{avgbg}62.20
& 56.03 & 31.25 & 72.06 & 60.44 & \cellcolor{avgbg}64.16 \\

\cellcolor{archbg}MemoryOS 
& 32.33 & 31.06 & 48.72 & 80.00 & 64.29 & 30.00 & \cellcolor{avgbg}44.66
& 56.74 & 45.83 & 67.06 & 40.19 & \cellcolor{avgbg}58.25 \\

\cellcolor{archbg}LightMem 
& 67.18 & 71.74 & \underline{83.12} & 87.14 & 32.14 & 68.18 & \cellcolor{avgbg}69.81
& 62.06 & 42.71 & 74.67 & 74.14 & \cellcolor{avgbg}70.26 \\

\cellcolor{archbg}SimpleMem
& 69.17 & 60.90 & 78.21 & 85.71 & 75.00 & 73.33 & \cellcolor{avgbg}71.60
& 64.50 & 44.90 & 76.80 & 74.50 & \cellcolor{avgbg}72.08 \\

\cellcolor{archbg}GAM 
& 60.15 & 70.68 & 78.21 & 75.71 & 94.64 & 70.00 & \cellcolor{avgbg}72.40
& 79.07 & 57.29 & 86.08 & 73.83 & \cellcolor{avgbg}80.45 \\

\cellcolor{promptbg}GEPA
& 69.17 & 71.43 & 79.49 & 88.57 & 78.57 & 76.67 & \cellcolor{avgbg}75.60
& 79.48 & \textbf{64.28} & 82.14 & 77.14 & \cellcolor{avgbg}79.50 \\

\cellcolor{promptbg}MetaMem
& 68.80 & 71.10 & 78.20 & 88.90 & 63.50 & \textbf{91.70} & \cellcolor{avgbg}74.47
& \multicolumn{4}{c}{--} & \cellcolor{avgbg}-- \\

\cellcolor{MemProbg}MemPro-5
& 73.86 & \underline{72.41} & 80.93 & 94.22 & 54.68 & 82.11 & \cellcolor{avgbg}75.77\basewin
& \underline{80.47} & 58.96 & 87.84 & 74.63 & \cellcolor{avgbg}81.94\basewin \\

\cellcolor{MemProbg}MemPro-10
& \underline{75.68} & 71.96 & 82.08 & \underline{96.31} & 58.24 & 83.72 & \cellcolor{avgbg}\underline{77.11}
& 79.82 & 60.37 & \underline{89.26} & \underline{77.54} & \cellcolor{avgbg}\underline{83.29} \\

\cellcolor{MemProbg}MemPro-15
& \textbf{77.44} & \textbf{73.68} & \textbf{83.33} & \textbf{98.57} & 60.71 & \underline{86.67} & \cellcolor{avgbg}\textbf{79.00}
& \textbf{82.26} & \underline{62.50} & \textbf{90.01} & \textbf{80.68} & \cellcolor{avgbg}\textbf{84.93} \\

\midrule

\rowcolor{gray!12}
\multicolumn{13}{l}{\textbf{\textit{Qwen3-30B-A3B-Instruct-2507}}} \\
\midrule

Full Text 
& 33.08 & 35.61 & 76.92 & 82.86 & 87.50 & 50.00 & \cellcolor{avgbg}54.67
& 69.86 & 57.29 & \textbf{87.40} & 51.71 & \cellcolor{avgbg}74.87 \\

RAG 
& 36.84 & 47.73 & 65.38 & 91.43 & \textbf{98.21} & 70.00 & \cellcolor{avgbg}60.69
& 62.41 & 57.29 & 76.81 & 47.98 & \cellcolor{avgbg}66.95 \\

\cellcolor{archbg}LangMem 
& 37.60 & 38.35 & 67.95 & 78.57 & 42.86 & 70.00 & \cellcolor{avgbg}50.80
& 53.22 & 50.01 & 63.54 & 30.13 & \cellcolor{avgbg}53.84 \\

\cellcolor{archbg}Mem0 
& 41.94 & 28.13 & 28.57 & 55.32 & 26.09 & \underline{81.82} & \cellcolor{avgbg}38.67
& 42.91 & 46.88 & 46.37 & 34.58 & \cellcolor{avgbg}43.31 \\

\cellcolor{archbg}A-MEM 
& 51.88 & 51.12 & 76.93 & 90.00 & \underline{96.43} & 40.00 & \cellcolor{avgbg}65.20
& 57.45 & 43.75 & 67.90 & 27.73 & \cellcolor{avgbg}56.10 \\

\cellcolor{archbg}MemoryOS 
& 28.57 & 36.84 & 61.54 & 72.86 & 92.86 & 33.33 & \cellcolor{avgbg}49.60
& 52.48 & 40.62 & 61.59 & 26.48 & \cellcolor{avgbg}51.30 \\

\cellcolor{archbg}LightMem 
& 54.20 & 51.91 & 66.67 & 80.00 & 31.25 & 80.00 & \cellcolor{avgbg}58.13
& 70.57 & 60.42 & 79.19 & 54.83 & \cellcolor{avgbg}71.36 \\

\cellcolor{archbg}SimpleMem
& 65.41 & 59.40 & 75.64 & 84.29 & 69.64 & 76.67 & \cellcolor{avgbg}69.20
& 72.00 & 63.00 & 80.10 & 58.90 & \cellcolor{avgbg}73.13 \\

\cellcolor{archbg}GAM 
& 67.67 & 67.29 & \textbf{82.05} & \underline{92.14} & 66.07 & 63.33 & \cellcolor{avgbg}72.80
& 71.98 & \textbf{75.00} & 80.61 & 60.43 & \cellcolor{avgbg}74.46 \\

\cellcolor{promptbg}GEPA
& 68.42 & 71.43 & \textbf{82.05} & 90.00 & 80.36 & 76.67 & \cellcolor{avgbg}76.20
& \underline{74.82} & 65.62 & 82.51 & 61.99 & \cellcolor{avgbg}75.77 \\

\cellcolor{promptbg}MetaMem
& \underline{69.60} & 69.24 & 79.18 & 90.70 & 38.14 & \textbf{94.16} & \cellcolor{avgbg}71.90
& \multicolumn{4}{c}{--} & \cellcolor{avgbg}-- \\

\cellcolor{MemProbg}MemPro-5
& 68.73 & 73.82 & 79.64 & 90.87 & 82.59 & 77.46 & \cellcolor{avgbg}76.96\basewin
& 74.36 & 67.94 & 82.20 & 65.18 & \cellcolor{avgbg}76.33\basewin \\

\cellcolor{MemProbg}MemPro-10
& 68.21 & \underline{75.03} & \underline{80.84} & 91.76 & 94.12 & 78.39 & \cellcolor{avgbg}\underline{78.80}
& 73.92 & 69.63 & 83.07 & \underline{66.43} & \cellcolor{avgbg}\underline{77.09} \\

\cellcolor{MemProbg}MemPro-15
& \textbf{71.43} & \textbf{75.94} & \textbf{82.05} & \textbf{92.86} & \textbf{98.21} & 80.00 & \cellcolor{avgbg}\textbf{80.80}
& \textbf{75.17} & \underline{70.83} & \underline{83.47} & \textbf{67.60} & \cellcolor{avgbg}\textbf{77.85} \\

\bottomrule
\end{tabular}
}

\vspace{2pt}
\begin{tabular}{lll}
\cellcolor{archbg}\strut Static agentic memory system &
\cellcolor{promptbg}\strut Prompt-level evolution &
\cellcolor{MemProbg}\strut System-level evolution
\end{tabular}

\caption{
Results on LongMemEval and LoCoMo. All reported results are LLM-as-a-Judge scores.
\textbf{Bold} and \underline{underline} indicate the best and second-best results per backbone.
Avg. denotes the mean across subcategories.
MemPro-5/10/15 denote MemPro after 5/10/15 evolution iterations.
$\dagger$ indicates that MemPro-5 already surpasses all non-MemPro baselines on the corresponding Avg. score.
}
\label{tab:longmemeval_locomo}
\vspace{-8pt}
\end{table*}

\section{Experiments}
\subsection{Experimental Setup}
\label{sec:exp_setup}
\paragraph{Datasets and splits.}
To comprehensively evaluate the effectiveness of MemPro, we conduct experiments on four datasets covering both agentic memory and multi-hop QA scenarios. The agentic memory datasets include LongMemEval~\citep{wu2024longmemeval} and LoCoMo~\citep{locomo}, while the multi-hop QA datasets include HotpotQA~\citep{hotpotqa} and NarrativeQA (full-document)~\citep{narrativeqa}. For LongMemEval and LoCoMo, we randomly sample 10\% of the examples from each dataset for MemPro training and use the remaining examples for testing. For HotpotQA, we consider three context-length settings---56K, 224K, and 448K---and randomly sample 30 questions (23\%) for training under each setting, using the remaining questions for testing. For NarrativeQA, we randomly sample 40 questions (12\%) for training and use the rest for testing. Detailed category breakdowns and additional dataset information are provided in Appendix~\ref{app:datasets}.

\paragraph{Metrics.}
For LongMemEval and LoCoMo, we use LLM-as-a-Judge for evaluation, with GPT-4o-mini~\cite{gpt4omini} as the judge model across all experiments. The judge prompts for LongMemEval and LoCoMo follow GAM~\cite{gam} and LightMem~\cite{lightmem}, respectively. For HotpotQA and NarrativeQA, we report word-level F1~\cite{gam} between the predicted answer and the gold answer. The judge prompts are provided in \autoref{fig:locomo_judge_prompt} and \autoref{fig:longmemeval_unified_judge_prompt}.

\paragraph{Baselines.}
We compare MemPro against a series of representative baselines: (1) Direct-context methods: Full Text and Naive RAG, which represent baselines without an agentic framework. (2) Static agentic memory systems: LangMem~\citep{langmem}, Mem0~\citep{mem0}, A-Mem~\citep{amem}, MemoryOS~\citep{memoryos}, LightMem~\citep{lightmem}, SimpleMem~\citep{simplemem}, and GAM~\citep{gam}, which represent strong existing agentic memory baselines. (3) Prompt-level evolution methods: GEPA~\citep{gepa} and MetaMem~\citep{metamem}, which implement prompt-level self-evolution on top of static memory systems. More details about the baselines are provided in Appendix~\ref{app:baselines}.

\paragraph{Implementation Details.}
During MemPro evolution, the \textit{Memory Agent} and \textit{Research Agent} in the MCR pipeline use gpt-4o-mini as the backbone, while the \textit{Evolving Agent} is implemented with a Codex harness using gpt-5.4-medium. We set the maximum outer evolution and inner expansion iterations to 15 and 20, respectively. For~\autoref{tab:longmemeval_locomo}, we report two evaluation-time backbone settings, \texttt{gpt-4o-mini} and \texttt{Qwen3-30B-A3B-Instruct-2507}; in each setting, the \textit{Memory Agent} and \textit{Research Agent} use the same backbone. Among the baselines, GEPA, SimpleMem, and GAM on LongMemEval are reproduced under settings aligned with MemPro, while the remaining results are taken from the corresponding papers~\citep{gam, lightmem, metamem}. For~\autoref{tab:qa_results}, all baseline results are taken from the corresponding papers~\citep{gam}. We replace the Qwen3-30B-A3B backbone with Qwen2.5-14B to match the baseline settings for fair comparison. Baselines missing from~\autoref{tab:qa_results} are omitted because they do not support HotpotQA or NarrativeQA. For~\autoref{fig:efficiency}, all baselines are reproduced under settings aligned with MemPro. Further implementation details are provided in Appendix~\ref{app:implementation}.
Notably, all results from our runs, including MemPro and reproduced baselines, are averaged over three runs.

\begin{table*}[t]
\centering
\scriptsize
\setlength{\tabcolsep}{4.2pt}
\renewcommand{\arraystretch}{1.08}
\resizebox{0.95\textwidth}{!}{
\begin{tabular}{lcccccccccc}
\toprule
\multirow{3}{*}{\textbf{Method}} 
& \multicolumn{5}{c}{\textbf{\textit{GPT-4o-mini}}} 
& \multicolumn{5}{c}{\textbf{\textit{Qwen2.5-14B}}} \\
\cmidrule(lr){2-6} \cmidrule(lr){7-11}
& \multicolumn{4}{c}{\textbf{HotpotQA}} 
& \multirow{2}{*}{\textbf{NarrativeQA}}
& \multicolumn{4}{c}{\textbf{HotpotQA}} 
& \multirow{2}{*}{\textbf{NarrativeQA}} \\
\cmidrule(lr){2-5} \cmidrule(lr){7-10}
& \textbf{56K} & \textbf{224K} & \textbf{448K} & \textbf{Avg.} & 
& \textbf{56K} & \textbf{224K} & \textbf{448K} & \textbf{Avg.} &  \\
\midrule

Full Text 
& 56.56 & 54.29 & 53.92 & 54.92 & 31.26
& 49.75 & 46.82 & 43.17 & 46.58 & 29.69 \\

RAG 
& 52.71 & 51.84 & 54.01 & 52.85 & 25.00
& 51.81 & 46.72 & 48.36 & 48.96 & 18.29 \\

A-MEM 
& 33.90 & 30.22 & 31.37 & 31.83 & 27.07
& 27.04 & 25.65 & 22.92 & 25.20 & 25.18 \\

Mem0 
& 32.58 & 31.74 & 27.41 & 30.58 & 29.16
& 30.12 & 32.44 & 26.55 & 29.70 & 27.80 \\

MemoryOS 
& 26.47 & 23.10 & 24.16 & 24.58 & 26.70
& 24.58 & 30.25 & 23.13 & 25.99 & 23.45 \\

LightMem 
& 40.93 & 35.28 & 30.02 & 35.41 & 17.51
& 37.30 & 27.72 & 28.25 & 31.09 & 16.57 \\

GAM 
& 63.22 & 61.46 & 59.81 & 61.50 & 36.86
& 64.07 & 55.99 & 57.87 & 59.31 & 30.17 \\

MemPro-5
& 65.18\basewin & 61.72\basewin & 60.84\basewin & 62.58\basewin & 37.23\basewin
& 64.36\basewin & 58.73\basewin & 57.91\basewin & 60.33\basewin & 30.34\basewin \\

MemPro-10
& \underline{68.05} & \underline{63.94} & \underline{62.36} & \underline{64.78} & \underline{37.84}
& \underline{65.41} & \underline{60.54} & \underline{58.76} & \underline{61.57} & \underline{30.78} \\

MemPro-15
& \textbf{70.32} & \textbf{65.81} & \textbf{64.02} & \textbf{66.72} & \textbf{38.12}
& \textbf{66.98} & \textbf{62.28} & \textbf{59.80} & \textbf{63.02} & \textbf{31.23} \\

\bottomrule
\end{tabular}
}
\caption{
Results on HotpotQA and NarrativeQA.
All reported results are F1 scores.
For HotpotQA, 56K/224K/448K denote context lengths, and Avg. denotes their mean.
MemPro-5/10/15 denote MemPro after 5/10/15 evolution iterations.
\textbf{Bold} and \underline{underline} indicate the best and second-best results per backbone.
$\dagger$ indicates that MemPro-5 already surpasses all non-MemPro baselines on the corresponding score.
}
\label{tab:qa_results}
\vspace{-8pt}
\end{table*}

\subsection{Performance on Memory Benchmarks}
To evaluate MemPro and the baselines in long-term interaction scenarios, we conduct experiments on LongMemEval and LoCoMo. \autoref{tab:longmemeval_locomo} shows the results, with key observations summarized below.

\paragraph{Naive Compression Hurts Performance, but Better Memory Design Can Overcome It.} As shown in \autoref{tab:longmemeval_locomo}, Full Text and Naive RAG outperform several early memory systems (LangMem, Mem0, MemoryOS), suggesting that naive compression can trade accuracy for context capacity. Yet recent systems (LightMem, SimpleMem, GAM) also compress memory while surpassing Full Text and Naive RAG, showing that this loss can be overcome by better memory-system design.

\paragraph{Prompt-Level Evolution Improves Agentic Memory Systems.} The results in \autoref{tab:longmemeval_locomo} show that prompt-level evolving memory systems achieve stronger overall performance than static memory systems. This demonstrates that even prompt-level evolution alone can effectively improve performance, highlighting the necessity of evolving components beyond the memory bank in agentic memory systems.

\paragraph{MemPro Achieves SOTA with Only a Few Evolution Iterations.}
As shown in \autoref{tab:longmemeval_locomo}, with only 5 evolution iterations, MemPro-5 consistently outperforms the strongest baseline on each benchmark and backbone LLM, achieving state-of-the-art (SOTA) performance. This shows that MemPro can achieve strong performance with a low evolution cost, demonstrating that evolving components beyond the memory bank and prompts can further improve performance. These results highlight the promise of treating the entire memory system as the target of evolution. While MemPro leads on overall averages, it underperforms several baselines on the Single-Assistant (\textbf{Asst.}) category under the gpt-4o-mini backbone; we analyze this category- and backbone-specific gap in Appendix~\ref{app:single_assistant}.

\paragraph{MemPro Continues to Improve as Evolution Progresses After Reaching SOTA.}
Although MemPro-5 already outperforms the strongest baseline on each benchmark and backbone LLM, MemPro continues to improve steadily as evolution progresses. As shown in \autoref{tab:longmemeval_locomo}, using the Avg. columns of each benchmark and averaging them across the two backbones, MemPro improves by +1.59 on LongMemEval and +1.06 on LoCoMo from MemPro-5 to MemPro-10, and further improves by +1.95 on LongMemEval and +1.20 on LoCoMo from MemPro-10 to MemPro-15. By the 15th evolution iteration, MemPro achieves superior performance, demonstrating the effectiveness and high ceiling of its evolution process. Figure~\ref{fig:evolution_dynamics} visualizes this evolution trajectory, and a qualitative case study is provided in Appendix~\ref{app:version_tree_case_study}.

\subsection{Performance on QA Benchmarks}
Beyond the memory benchmarks, we also evaluate MemPro and the baselines on HotpotQA and NarrativeQA to assess whether long-context memory capabilities transfer to knowledge-intensive QA scenarios. \autoref{tab:qa_results} reports the results, from which we draw observations broadly consistent with those on the memory benchmarks.
\paragraph{Direct Context Methods Beat Early Memory Systems but Trail Strong Ones.} Consistent with the memory benchmarks, direct-context methods beat early memory systems but trail strong ones such as GAM, while MemPro leads.

\paragraph{MemPro Surpasses Baselines Within a Few Iterations and Keeps Improving.} The results in~\autoref{tab:qa_results} show that MemPro surpasses all non-MemPro baselines as early as the fifth evolution iteration across benchmark settings and backbone models. After reaching this strong level, MemPro continues to improve with evolution.

\noindent Experiments on the multi-hop QA benchmarks further corroborate our observations and demonstrate the robustness of MemPro’s performance.

\begin{figure}[h]
\centering
\includegraphics[width=\columnwidth]{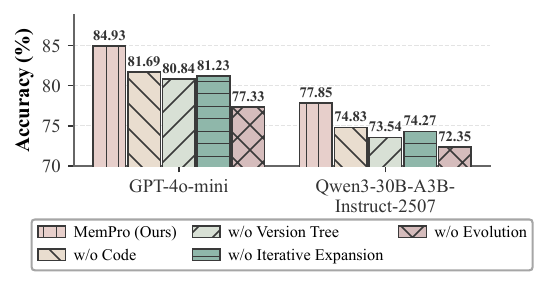}
\caption{Ablation study on the LoCoMo.}
\label{fig:ablation}
\end{figure}

\subsection{Ablation Study}
We conduct ablations on LoCoMo for four key designs:
(1) \textit{w/o Code} removes code-level edits, reducing MemPro to prompt-level evolution;
(2) \textit{w/o Version Tree} degenerates tree evolution into chain evolution, expanding only from the latest version;
(3) \textit{w/o Evolution} keeps only one outer evolution iteration;
and (4) \textit{w/o Iterative Expansion} uses a single edit during expansion without inner iterations. \autoref{fig:ablation} supports the following observations.

\paragraph{All components contribute to MemPro.}
Removing any key component leads to clear performance drops, showing that MemPro's gains come from system-level edits, tree evolution, multi-round evolution, and iterative expansion.

\paragraph{System-Level Evolution Matters.}
The \textit{w/o Code} variant reduces MemPro to prompt-level evolution and clearly underperforms the full MemPro. This supports our central claim that self-evolution should go beyond prompt text: editing executable pipeline components expands the optimization space and strengthens system-level evolution.
\paragraph{Tree Evolution Outperforms Chain Evolution.}
The \textit{w/o Version Tree} variant degenerates tree evolution into chain evolution and performs worse than MemPro. This shows that maintaining a version tree is more effective than following a single linear trajectory, as it allows MemPro to expand from strong historical versions rather than being constrained to the latest version.

\paragraph{Outer Evolution Enables Continuous Improvement.}
The \textit{w/o Evolution} variant keeps only one outer evolution iteration and degrades performance. This demonstrates that multi-round evolution over the version tree is important for continuous pipeline improvement, allowing later iterations to build on evaluated versions and discover better versions.
\paragraph{Iterative Expansion Matters.}
The \textit{w/o Iterative Expansion} variant performs only a single edit during expansion and also degrades performance. This indicates that expansion should not be treated as a one-shot edit; inner expansion iterations allow the \textit{Evolving Agent} to refine the selected pipeline using accumulated diagnostic analysis before finalizing the new version.

We also provide a retrieval-tool ablation in Appendix~\ref{app:retrieval_ablation}, showing that BM25, embedding, and PAGE-ID retrieval are complementary.

\subsection{Efficiency Analysis}
\begin{figure}[h]
\centering
\includegraphics[width=\columnwidth]{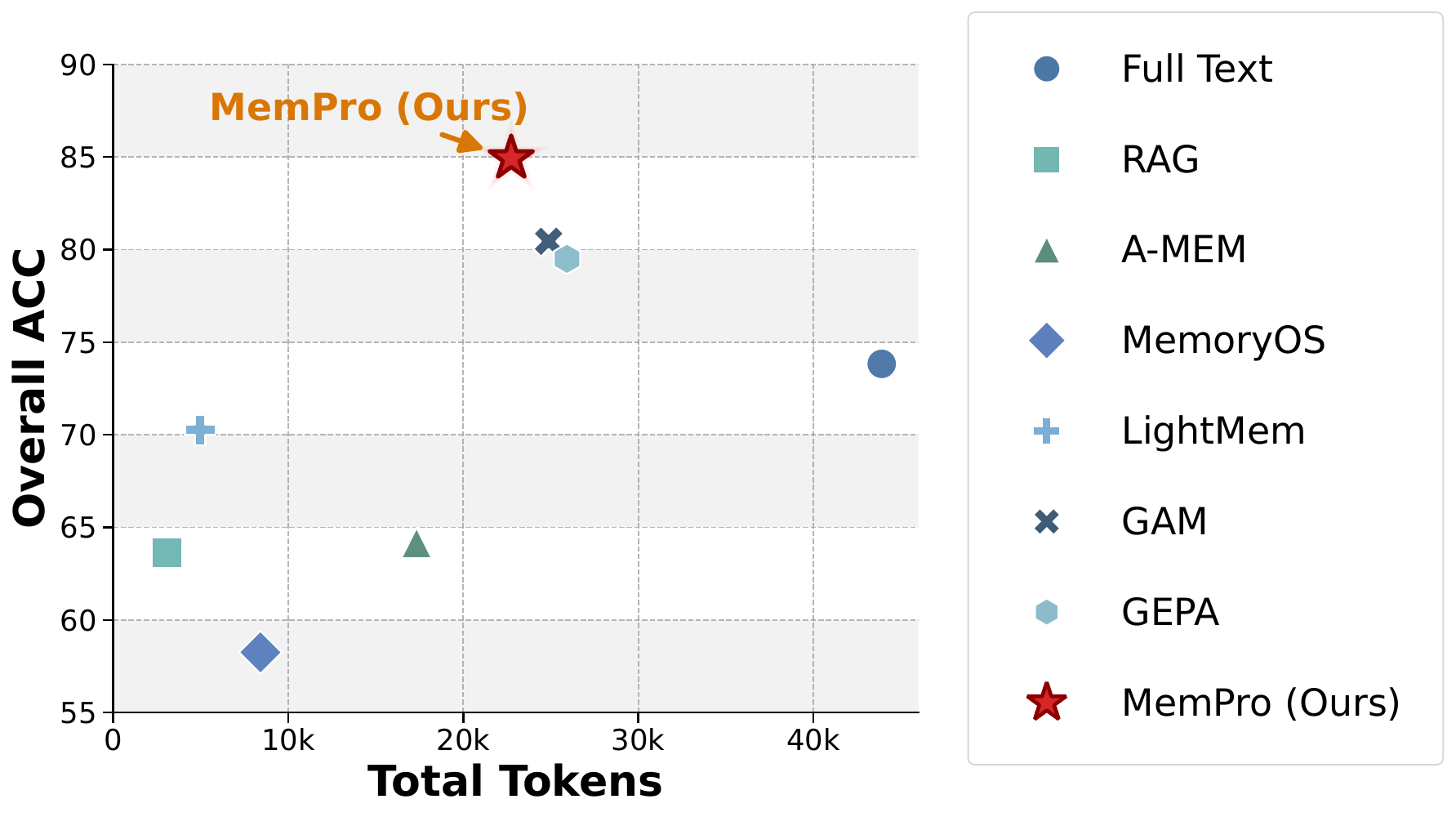}
\caption{Efficiency analysis on LoCoMo.}
\label{fig:efficiency}
\vspace{-10pt}
\end{figure}

We conduct an efficiency analysis on LoCoMo to compare accuracy and token cost, with results shown in \autoref{fig:efficiency}. MemPro achieves the highest overall accuracy with a reasonable token cost. It substantially outperforms Full Text while using fewer tokens, and surpasses strong memory baselines such as GAM and GEPA with comparable token budgets. Although lightweight methods are cheaper, they show much lower accuracy. Overall, MemPro offers a favorable trade-off between memory performance and token efficiency.

\section{Conclusion}
We presented \textbf{MemPro}, a system-level evolution framework that treats the entire memory construction--retrieval pipeline as an evolvable program rather than adapting only the memory bank or prompt text. Motivated by the task heterogeneity and memory--pipeline misalignment of fixed-pipeline memory systems, MemPro maintains a version tree of runnable pipeline implementations and, through iterative selection, expansion, and evaluation, uses an \textit{Evolving Agent} with failure-mode-guided edit--debug refinement to evolve both prompts and executable code. Across LongMemEval, LoCoMo, HotpotQA, and NarrativeQA, MemPro consistently surpasses strong static and prompt-level evolving baselines within a few iterations, keeps improving with evolution, and attains a favorable performance--cost trade-off, suggesting that agentic memory should be optimized as a runnable, self-evolving system.

\section{Limitations}

MemPro introduces an offline evolution stage in addition to task-time inference. This cost can be amortized when the evolved memory system is reused across many queries, and future work could further improve the efficiency of the evolution process. In addition, our implementation uses a capable \textit{Evolving Agent} to edit and debug runnable MCR pipelines; future work could study more efficient or specialized evolving agents. Finally, MemPro currently uses a tree-structured evolution framework, where each new pipeline version is derived from a single parent version. We have not explored more general evolution topologies, such as graph-structured evolution that allows multiple versions to be merged or recombined. Future work could study whether these alternatives improve the diversity and efficiency of system-level evolution.
\section{Ethical Considerations}

MemPro targets agentic memory systems that retain and reuse historical information. In real-world applications, such memories may contain sensitive user preferences, interaction histories, personal information, or task-specific records. Responsible deployment should therefore include privacy and data-governance practices such as user consent, data minimization, access control, and mechanisms for users to inspect, correct, or delete stored memories. Our experiments use public benchmark datasets and do not collect private user data.

Because MemPro edits executable components of the memory construction--retrieval pipeline, evolved versions should be sandboxed, logged, and reviewed before deployment in user-facing or safety-critical settings. In our experiments, evolution is restricted to the MCR pipeline, while evaluation data, gold answers, judge prompts, and held-out examples are not editable or exposed to the \textit{Evolving Agent}.

\bibliography{custom}

\appendix

\section{Appendix}
\label{app:appendix}

\subsection{Case Study}
\label{app:version_tree_case_study}

\autoref{fig:evolution_dynamics} shows how MemPro gradually improves the LoCoMo pipeline through the version tree. Starting from the initial framework, MemPro first improves ACC from 77.33 to 79.10 by refining temporal answers. This edit makes the model answer time-related questions with more direct and specific expressions, such as dates, months, or time spans, instead of producing broad explanations. The next major improvement comes from a question-type-aware integration strategy, which raises ACC to 80.88. This code-level change lets the system merge retrieved memories differently for different question types, so that temporal, counting, entity-centric, and multi-hop questions can preserve the information they need. Then, count and duration reasoning further improves ACC to 82.12 by explicitly handling questions that require counting repeated events or computing temporal spans. After that, adaptive retrieval depth increases ACC to 83.46 by searching more when the question needs broader evidence and avoiding unnecessary noisy retrieval for simpler questions. Finally, Focused evidence snippets improves ACC to 84.93 by placing salient evidence before integration, making key snippets easier for the model to use and reducing interference from long-context noise. Overall, MemPro improves ACC by 7.60 percentage points over the initial framework. The curve also shows several non-monotonic intermediate versions, which means that some edits introduce regressions. The version tree mitigates this problem by preserving strong previous versions and allowing later iterations to branch from them, rather than forcing evolution to follow a single linear path.

\subsection{Per-Category Analysis on LongMemEval: Single-Assistant}
\label{app:single_assistant}
Across most LongMemEval categories MemPro improves steadily, but on the
Single-Assistant (\textbf{Asst.}) category under the gpt-4o-mini
backbone it underperforms several baselines: MemPro-15 reaches 60.71, whereas
direct-context methods and strong static systems such as RAG (98.21),
A-MEM (96.43), and GAM (94.64) score much higher. We note three points that
place this gap in context.

First, the gap is backbone-specific rather than fundamental. With the
Qwen3-30B-A3B-Instruct-2507 backbone, the same evolved pipeline attains
98.21 on Single-Assistant, matching the best baseline. This suggests that the
limitation arises from the interaction between this category and the weaker
gpt-4o-mini backbone, not from system-level evolution itself.

Second, Single-Assistant questions ask about the assistant's own prior
statements, whose answers often depend on the assistant's specific earlier
wording. The MCR pipeline stores compressed, abstractive memories rather than
raw turns, so assistant-side phrasing can be lost during memory
construction---precisely the failure mode where uncompressed methods such as
Full Text and RAG retain an advantage. Under a weaker backbone, the
\textit{Memory Agent}'s abstraction tends to be more lossy, amplifying this
effect.

Third, the evolution signal for this category is sparse. Single-Assistant
contains 56 questions in LongMemEval, of which only about six (10\%) are sampled
into the evolution set. The \textit{Evolving Agent} therefore receives very few
failure cases for this category, and because selection and evaluation are driven
by the overall score, optimization naturally concentrates on more frequent
failure modes.

Taken together, these observations indicate that the gap reflects a
category- and backbone-specific trade-off rather than a defect of system-level
evolution. A natural remedy is to make evolution category-aware---for example,
adding a category-balanced objective or preserving raw assistant-side snippets
alongside abstractive memories---which we leave to future work.

\subsection{Dataset Details}
\label{app:datasets}

LoCoMo~\citep{locomo} is a long-term conversational memory benchmark built
from multi-session user--assistant histories. It contains four question types:
Single Hop, Multi Hop, Temporal, and Open Domain. LoCoMo contains 1,540
questions in total. We randomly sample 154 questions, corresponding to 10\% of
the full set, for framework evolution, and reserve the remaining 1,386
questions as a question-disjoint held-out evaluation set. We follow the GAM
GitHub evaluation protocol and judge prompt, and use \texttt{gpt-4o-mini} as
the judge model.

LongMemEval~\citep{wu2024longmemeval} evaluates long-term memory over
multi-session interactions. It covers six categories: Temporal Reasoning,
Multi Session, Knowledge Update, Single User, Single Assistant, and Single
Preference. The benchmark contains 500 questions. We randomly sample 50
questions for framework evolution and evaluate on the remaining 450
question-disjoint held-out questions. We use an LLM-as-judge evaluator with
\texttt{gpt-4o-mini} as the judge model; the judge prompt is provided in
\autoref{fig:locomo_judge_prompt} and \autoref{fig:longmemeval_unified_judge_prompt}.

HotpotQA~\citep{hotpotqa} is a multi-hop question-answering benchmark. We use
the dataset-provided long-context settings with approximately 56K, 224K, and
448K tokens. These settings use the same underlying questions but provide
different retrieved contexts. For each context length, we use the same 128
questions. The same 30 question IDs are used for evolution across all three
context lengths, and the remaining 98 questions are used for held-out
evaluation at each length. The 30 training question IDs across the three
context lengths form 90 question-context instances, and a single HotpotQA
framework is evolved on this mixed training set.

NarrativeQA~\citep{narrativeqa} is a narrative reading-comprehension benchmark
over long stories. We use the full-document setting rather than the summary
setting: each document is chunked, converted into memory, retrieved at question
time, and used for answer generation. We randomly sample 40 examples for
framework evolution and report final results on a disjoint 300-example
held-out evaluation set.

\subsection{Baseline Details}
\label{app:baselines}

We compare MemPro with direct-context methods, fixed memory-architecture
systems, and optimization-based memory methods.

Full Text feeds the complete available history or document directly to the
inference model. RAG retrieves query-relevant chunks from the available context
and answers from the retrieved evidence.

LangMem, Mem0~\citep{mem0}, A-MEM~\citep{amem},
MemoryOS~\citep{memoryos}, LightMem~\citep{lightmem}, and GAM~\citep{gam}
are fixed memory architectures. These systems maintain fixed memory-writing
and memory-reading procedures at evaluation time. On LoCoMo and LongMemEval,
we reproduce GAM and evaluate it under the same held-out protocol as MemPro.

GEPA~\citep{gepa} is a prompt-only optimization baseline. We run GEPA with the
same evolution subsets, optimizer model, and iteration budget as MemPro. GEPA
can edit framework prompts but cannot modify framework code or control flow.
Its final version is selected on the evolution subset and then evaluated on
the same held-out examples as MemPro.

MemPro without evolution denotes our manually initialized framework before
failure-driven evolution, corresponding to \texttt{v0000} in the version tree.
MetaMem results are taken from the original paper where published results are
available.

MemPro, MemPro without evolution, GEPA, and the reproduced GAM results are
evaluated under our held-out protocol. Other previously published memory
baselines are reported from LightMem or GAM under the matched benchmark and
backbone setting, and MetaMem results are taken from the original paper.

\subsection{Implementation Details}
\label{app:implementation}

MemPro evolves runnable MCR framework versions rather than isolated prompt
strings. Each version contains both the task-facing prompts and the executable
code used by the \textit{Memory Agent} and \textit{Research Agent}, including
memory construction, retrieval, evidence integration, context construction,
and final-answer generation. The \textit{Evolving Agent} is implemented with
the OpenAI Codex coding harness using \texttt{gpt-5.4-medium}. It takes
\texttt{AGENTS.md} as the task-level instruction file, which specifies the
procedures for base-version selection, edit--debug iteration, and training-set
performance analysis.

During evolution, all task-time MCR reasoning calls are performed with
\texttt{gpt-4o-mini} using temperature $0.0$ and top-$p$ $1.0$. The initial
retrieval module uses three retrieval channels: BM25 keyword retrieval,
\texttt{BAAI/bge-m3} semantic retrieval, and PAGE-ID retrieval. Each channel
returns the top-5 candidates by default. For each question, the
\textit{Research Agent} performs at most five retrieval--integration
iterations. The maximum evolution budget is capped at 15 iterations.

GEPA is given the same training set and iteration budget as MemPro, and uses
the same optimizer model \texttt{gpt-5.4-medium}. However, GEPA is restricted to
prompt edits and cannot modify executable framework code or control flow. For
model-transfer experiments, we reuse the framework version evolved under
\texttt{gpt-4o-mini} and replace only the task-time inference model with
\texttt{Qwen3-30B-A3B-Instruct-2507}, without any additional evolution.
\texttt{Qwen3-30B-A3B-Instruct-2507} is served with SGLang on eight H800 GPUs
using bfloat16 precision, and decoded with temperature $0.7$ and top-$p$
$0.9$. All LLM-as-a-Judge evaluations use \texttt{gpt-4o-mini} with
temperature $0$. Token cost is computed over task-time inference only and
excludes offline evolution calls.

\subsection{Retrieval Ablation Study}
\label{app:retrieval_ablation}

\begin{figure}[t!]
\centering
\includegraphics[width=\columnwidth]{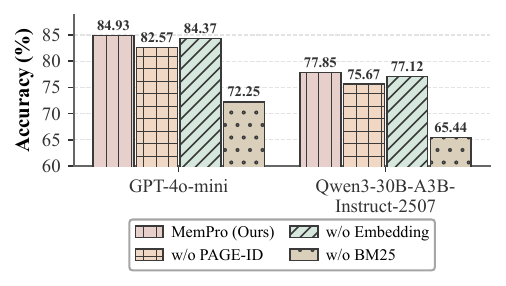}
\caption{Retrieval ablation study on the LoCoMo. Results are reported with \texttt{gpt-4o-mini} and \texttt{Qwen3-30B-A3B-Instruct-2507}.}
\label{fig:retrieval_ablation}
\end{figure}

We analyze the contribution of different retrieval tools in MemPro on the LoCoMo held-out set. As shown in~\autoref{fig:retrieval_ablation}, removing any retrieval tool decreases performance under both backbone models, showing that keyword-based, semantic, and structural retrieval signals are complementary. BM25 is the most important retrieval channel: removing it drops accuracy from 84.93 to 72.25 with \texttt{gpt-4o-mini} and from 77.85 to 65.44 with \texttt{Qwen3-30B-A3B-Instruct-2507}. This suggests that exact keyword matching is crucial for long-term conversational memory, where answers often depend on specific names, events, dates, or surface-form cues. Embedding-based retrieval also contributes consistently: removing it drops accuracy from 84.93 to 82.57 with \texttt{gpt-4o-mini} and from 77.85 to 75.67 with \texttt{Qwen3-30B-A3B-Instruct-2507}. This indicates that semantic retrieval helps recover relevant memories when the query and stored memories use different wording. PAGE-ID retrieval has a smaller but still positive effect: removing it drops accuracy from 84.93 to 84.37 with \texttt{gpt-4o-mini} and from 77.85 to 77.12 with \texttt{Qwen3-30B-A3B-Instruct-2507}, suggesting that structural lookup provides useful auxiliary grounding. Overall, the best performance is achieved when MemPro coordinates all three retrieval tools, combining BM25 for exact keyword matching, embedding retrieval for semantic matching, and PAGE-ID retrieval for structural localization.

\begin{figure*}[t]
\centering
\begin{tcolorbox}[
  enhanced,
  width=0.95\textwidth,
  colback=orange!3,
  colframe=orange!65!black,
  colbacktitle=orange!80!brown,
  coltitle=white,
  fonttitle=\bfseries\large,
  title=Memory Construction Prompt,
  boxrule=1.0pt,
  arc=2mm,
  left=4mm,
  right=4mm,
  top=3mm,
  bottom=3mm,
  titlerule=0pt,
  fontupper=\small\sffamily
]
\setlength{\parindent}{0pt}
\setlength{\parskip}{0.6em}
\linespread{1.12}\selectfont

You are the MemoryAgent. Your job is to write one concise abstract that can be stored as long-term memory.

\textbf{MAIN OBJECTIVE:}

Generate a concise, self-contained and coherent abstract of INPUT\_MESSAGE that preserves ALL important information in INPUT\_MESSAGE.
MEMORY\_CONTEXT is provided so you can understand the broader situation such as people, modules, decisions, ongoing tasks and keep wording consistent.

\textbf{INPUTS:}

\textbf{MEMORY\_CONTEXT:}

\{memory\_context\}

\textbf{INPUT\_MESSAGE:}

\{input\_message\}

\textbf{YOUR TASK:}

1. Read INPUT\_MESSAGE and extract all specific, memory-relevant information, such as:

\quad - plans, goals, decisions, requests, preferences

\quad - actions taken, next steps, assignments, and responsibilities

\quad - problems, blockers, bugs, questions that need follow-up

\quad - specific facts such as names, dates, numbers, locations

2. Use MEMORY\_CONTEXT to:

\quad - resolve or disambiguate the entities, components, tasks, or resources mentioned in INPUT\_MESSAGE,

\quad - keep terminology, such as names of agents, modules, datasets, etc., consistent with prior usage,

\quad - include minimal background context if it is required for the abstract to be understandable.

\quad You MUST NOT invent or add information that appears only in MEMORY\_CONTEXT and is NOT implied or mentioned in INPUT\_MESSAGE.

3. Your abstract MUST:

\quad - summarize all important content from INPUT\_MESSAGE,

\quad - be understandable on its own without seeing INPUT\_MESSAGE,

\quad - be factual and specific.

\textbf{STYLE RULES:}

\quad - Output exactly ONE concise paragraph. No bullet points.

\quad - Do NOT include meta phrases like ``The user said...'' or ``The conversation is about...''.

\quad - Do NOT give advice, opinions, or suggestions.

\quad - Do NOT ask questions.

\quad - Do NOT include anything that is not grounded in INPUT\_MESSAGE.

\textbf{OUTPUT FORMAT:}

Return ONLY the single paragraph. Do NOT add any headings or labels.

\end{tcolorbox}

\caption{Memory construction prompt used by the \textit{Memory Agent} in the MCR pipeline to generate memory updates from structured segments and the previous memory bank.}
\label{fig:memoryagent_prompt}
\end{figure*}

\begin{figure*}[t]
\centering
\begin{tcolorbox}[
  enhanced,
  width=0.95\textwidth,
  colback=orange!3,
  colframe=orange!65!black,
  colbacktitle=orange!80!brown,
  coltitle=white,
  fonttitle=\bfseries\large,
  title=Retrieval Prompt,
  boxrule=1.0pt,
  arc=2mm,
  left=4mm,
  right=4mm,
  top=3mm,
  bottom=3mm,
  titlerule=0pt,
  fontupper=\small\sffamily
]
\setlength{\parindent}{0pt}
\setlength{\parskip}{0.6em}
\linespread{1.12}\selectfont

You are the Retrieval Agent. Make a retrieval plan for the question using the memory below.

\textbf{QUESTION:}

\{request\}

\textbf{MEMORY:}

\{memory\}

Return one JSON object with exactly these keys:

\quad - ``info\_needs'': array of missing facts or sub-questions

\quad - ``tools'': array using only [``keyword'', ``vector'', ``page\_index'']

\quad - ``keyword\_collection'': array of short keyword queries

\quad - ``vector\_queries'': array of short natural-language queries

\quad - ``page\_index'': array of integers, or []

If a field is not needed, use an empty array.

Return only JSON.

\end{tcolorbox}

\caption{Retrieval prompt for the \textit{Research Agent} in the MCR pipeline.}
\label{fig:retrieval_prompt}
\end{figure*}

\begin{figure*}[t]
\centering
\begin{tcolorbox}[
  enhanced,
  width=0.95\textwidth,
  colback=orange!3,
  colframe=orange!65!black,
  colbacktitle=orange!80!brown,
  coltitle=white,
  fonttitle=\bfseries\large,
  title=Integration Prompt,
  boxrule=1.0pt,
  arc=2mm,
  left=4mm,
  right=4mm,
  top=3mm,
  bottom=3mm,
  titlerule=0pt,
  fontupper=\small\sffamily
]
\setlength{\parindent}{0pt}
\setlength{\parskip}{0.6em}
\linespread{1.12}\selectfont

You are the Integration Agent. Merge the current result with the new evidence into one factual summary.

\textbf{QUESTION:}

\{question\}

\textbf{EVIDENCE\_CONTEXT:}

\{evidence\_context\}

\textbf{RESULT:}

\{result\}

Return one JSON object with exactly these keys:

\quad - ``content'': a factual summary

\quad - ``sources'': page ids that support the content

If there is no useful information, set ``content'' to ``'' and ``sources'' to [].

Return only JSON.

\end{tcolorbox}

\caption{Integration prompt for the \textit{Research Agent} in the MCR pipeline.}
\label{fig:integration_prompt}
\end{figure*}

\begin{figure*}[t]
\centering
\begin{tcolorbox}[
  enhanced,
  width=0.95\textwidth,
  colback=orange!3,
  colframe=orange!65!black,
  colbacktitle=orange!80!brown,
  coltitle=white,
  fonttitle=\bfseries\large,
  title=Reflection Prompt,
  boxrule=1.0pt,
  arc=2mm,
  left=4mm,
  right=4mm,
  top=3mm,
  bottom=3mm,
  titlerule=0pt,
  fontupper=\small\sffamily
]
\setlength{\parindent}{0pt}
\setlength{\parskip}{0.6em}
\linespread{1.12}\selectfont

You are the Reflection Agent. Decide whether the current result is enough to answer the question.

\textbf{QUESTION:}

\{question\}

\textbf{RESULT:}

\{result\}

Return one JSON object with exactly these keys:

\quad - ``enough'': true or false

\quad - ``next\_question'': if ``enough'' is false, generate the next question needed to obtain the missing information; otherwise return null

Return only JSON.

\end{tcolorbox}

\caption{Reflection prompt for the \textit{Research Agent} in the MCR pipeline.}
\label{fig:reflection_prompt}
\end{figure*}

\begin{figure*}[t]
\centering
\begin{tcolorbox}[
  enhanced,
  width=0.95\textwidth,
  colback=orange!3,
  colframe=orange!65!black,
  colbacktitle=orange!80!brown,
  coltitle=white,
  fonttitle=\bfseries\large,
  title=LoCoMo Judge Prompt,
  boxrule=1.0pt,
  arc=2mm,
  left=4mm,
  right=4mm,
  top=3mm,
  bottom=3mm,
  titlerule=0pt,
  fontupper=\small\sffamily
]
\setlength{\parindent}{0pt}
\setlength{\parskip}{0.6em}
\linespread{1.12}\selectfont

Your task is to label an answer to a question as ``CORRECT'' or ``WRONG''. You will be given the following data: (1) a question (posed by one user to another user), (2) a `gold' (ground truth) answer, (3) a generated answer which you will score as CORRECT/WRONG.

The point of the question is to ask about something one user should know about the other user based on their prior conversations. The gold answer will usually be a concise and short answer that includes the referenced topic, for example: Question: Do you remember what I got the last time I went to Hawaii? Gold answer: A shell necklace The generated answer might be much longer, but you should be generous with your grading - as long as it touches on the same topic as the gold answer, it should be counted as CORRECT.

For time related questions, the gold answer will be a specific date, month, year, etc. The generated answer might be much longer or use relative time references (like `last Tuesday' or `next month'), but you should be generous with your grading - as long as it refers to the same date or time period as the gold answer, it should be counted as CORRECT. Even if the format differs (e.g., `May 7th' vs `7 May'), consider it CORRECT if it's the same date.

Now it's time for the real question:

\textbf{Question:}

\{question\}

\textbf{Gold answer:}

\{gold\_answer\}

\textbf{Generated answer:}

\{generated\_answer\}

Return the label CORRECT or WRONG in a json format with the key as ``label''. Do NOT include both CORRECT and WRONG in your response, or it will break the evaluation script.

\end{tcolorbox}

\caption{Judge prompt used for LoCoMo evaluation.}
\label{fig:locomo_judge_prompt}
\end{figure*}

\begin{figure*}[t]
\centering
\begin{tcolorbox}[
  enhanced,
  width=0.95\textwidth,
  colback=orange!3,
  colframe=orange!65!black,
  colbacktitle=orange!80!brown,
  coltitle=white,
  fonttitle=\bfseries\large,
  title=LongMemEval Judge Prompt,
  boxrule=1.0pt,
  arc=2mm,
  left=4mm,
  right=4mm,
  top=3mm,
  bottom=3mm,
  titlerule=0pt,
  fontupper=\small\sffamily
]
\setlength{\parindent}{0pt}
\setlength{\parskip}{0.6em}
\linespread{1.12}\selectfont

I will give you a task type, a question, a correct answer, and a response from a model. Please answer yes if the response contains the correct answer. Otherwise, answer no.

If the response is equivalent to the correct answer or contains all the intermediate steps to get the correct answer, you should also answer yes. If the response only contains a subset of the information required by the answer, answer no.

For task types \texttt{single-session-user}, \texttt{single-session-assistant}, and \texttt{multi-session}, use the general rule above.

For task type \texttt{temporal-reasoning}, in addition to the general rule, do not penalize off-by-one errors for the number of days. If the question asks for the number of days/weeks/months, etc., and the model makes off-by-one errors (e.g., predicting 19 days when the answer is 18), the model's response is still correct.

For task type \texttt{knowledge-update}, if the response contains some previous information along with an updated answer, the response should be considered as correct as long as the updated answer is the required answer.

For task type \texttt{single-session-preference}, I will give you a question, a rubric for desired personalized response, and a response from a model. Please answer yes if the response satisfies the desired response. Otherwise, answer no. The model does not need to reflect all the points in the rubric. The response is correct as long as it recalls and utilizes the user's personal information correctly.

\textbf{Task type:}

\{task\}

\textbf{Question:}

\{question\}

\textbf{Correct Answer / Rubric:}

\{answer\}

\textbf{Model Response:}

\{response\}

Is the model response correct? Answer yes or no only.

\end{tcolorbox}

\caption{Judge prompt used for LongMemEval evaluation.}
\label{fig:longmemeval_unified_judge_prompt}
\end{figure*}

\begin{figure*}[t]
\centering
\begin{tcolorbox}[
  enhanced,
  width=0.95\textwidth,
  colback=orange!3,
  colframe=orange!65!black,
  colbacktitle=orange!80!brown,
  coltitle=white,
  fonttitle=\bfseries\large,
  title=Selection Prompt,
  boxrule=1.0pt,
  arc=2mm,
  left=4mm,
  right=4mm,
  top=3mm,
  bottom=3mm,
  titlerule=0pt,
  fontupper=\small\sffamily
]
\setlength{\parindent}{0pt}
\setlength{\parskip}{0.6em}
\linespread{1.12}\selectfont

You are an autonomous evolution agent for a memory-augmented question answering system.

Your task is to analyze previous evolution logs, select the most promising base version, and create a new editable child version from it.

\textbf{AVAILABLE TOOLS:}

\begin{itemize}[leftmargin=1.5em, itemsep=0.2em, topsep=0.2em]
    \item \texttt{\{Tool LogReader: read version records, metrics, parent-child relations, branch history, and analysis notes.\}}
    \item \texttt{\{Tool TraceInspector: inspect representative weak or regressed cases, including retrieved evidence, memory outputs, predictions, and traces.\}}
    \item \texttt{\{Tool BaseSelector: compare candidate base versions using historical performance, stability, and branch context.\}}
    \item \texttt{\{Tool VersionCreator: create a new editable child version from the selected base version.\}}
\end{itemize}

\textbf{SELECTION CRITERIA:}

\begin{itemize}[leftmargin=1.5em, itemsep=0.2em, topsep=0.2em]
    \item Prefer versions with strong overall training performance and high success rate.
    \item Prefer versions that improve over their parent while avoiding broad regressions.
    \item Prefer versions that improve recurring weak cases rather than isolated examples.
    \item Prefer versions with stable behavior across examples, categories, or question types.
    \item When scores are similar, prefer clearer failure analysis, lower regression risk, and better compatibility with the next improvement direction.
\end{itemize}

\textbf{OUTPUT FORMAT:}

Write a concise selection note covering the selected base version, its parent and branch position, the main reason for selection, key risks, recommended next improvement direction, whether memory artifacts can be reused, and the newly created child version for the edit stage.

Do not use held-out evaluation information, gold-answer shortcuts, sample-specific rules, or manual fixes tied to individual examples.

\end{tcolorbox}

\caption{Selection-stage prompt used by the \textit{Evolving Agent} to choose a promising base version from the MCR version tree based on node evaluation logs.
}
\label{app:selection_prompt}
\end{figure*}

\begin{figure*}[t]
\centering
\begin{tcolorbox}[
  enhanced,
  width=0.95\textwidth,
  colback=orange!3,
  colframe=orange!65!black,
  colbacktitle=orange!80!brown,
  coltitle=white,
  fonttitle=\bfseries\large,
  title=Expansion Prompt,
  boxrule=1.0pt,
  arc=2mm,
  left=4mm,
  right=4mm,
  top=3mm,
  bottom=3mm,
  titlerule=0pt,
  fontupper=\small\sffamily
]
\setlength{\parindent}{0pt}
\setlength{\parskip}{0.6em}
\linespread{1.12}\selectfont

You are an autonomous evolution agent improving a memory-augmented question answering system.

Your task is to edit the newly created child version, apply one coherent change direction, and refine it through targeted debugging before full training-set evaluation.

\textbf{AVAILABLE TOOLS:}

\begin{itemize}[leftmargin=1.5em, itemsep=0.2em, topsep=0.2em]
    \item \texttt{\{Tool CodeEditor: modify prompts or executable framework logic inside the current child version.\}}
    \item \texttt{\{Tool DebugRunner: run targeted diagnostic examples and return predictions, memory states, retrieved evidence, assembled context, and traces.\}}
    \item \texttt{\{Tool TraceInspector: compare the intended behavior with the actual trace and identify recurring failure patterns.\}}
    \item \texttt{\{Tool MemoryManager: decide whether previous memory artifacts can be reused or must be rebuilt.\}}
\end{itemize}

\textbf{EDIT--DEBUG CRITERIA:}

\begin{itemize}[leftmargin=1.5em, itemsep=0.2em, topsep=0.2em]
    \item Use exactly one coherent edit direction in each child version.
    \item The edit may target memory construction, retrieval planning, evidence integration, reflection, context assembly, or final-answer formatting.
    \item Debug on representative weak or regressed cases to identify recurring failure patterns.
    \item Revise the child version only when traces reveal a generalizable problem.
    \item Avoid fixes based on sample IDs, gold answers, or individual examples.
\end{itemize}

\textbf{MEMORY REUSE:}

Reuse memory artifacts only if the edit does not affect memory construction. Rebuild memory when the edit changes memory prompts, memory-writing logic, chunking, page construction, or any component that can alter stored memory content.

\textbf{OUTPUT FORMAT:}

Report the child version, base version, parent relationship, edit direction, actual changes made, memory reuse decision, targeted failure pattern, observed improvements and regressions in debugging, and whether the version is ready for full training-set evaluation.

\end{tcolorbox}

\caption{Expansion-stage prompt used by the \textit{Evolving Agent} to expand the selected MCR version into a new version through edit--debug refinement.}

\label{app:exp_prompt}
\end{figure*}

\begin{figure*}[t]
\centering
\begin{tcolorbox}[
  enhanced,
  width=0.95\textwidth,
  colback=orange!3,
  colframe=orange!65!black,
  colbacktitle=orange!80!brown,
  coltitle=white,
  fonttitle=\bfseries\large,
  title=Evaluation Prompt,
  boxrule=1.0pt,
  arc=2mm,
  left=4mm,
  right=4mm,
  top=3mm,
  bottom=3mm,
  titlerule=0pt,
  fontupper=\small\sffamily
]
\setlength{\parindent}{0pt}
\setlength{\parskip}{0.6em}
\linespread{1.12}\selectfont

You are an autonomous evolution agent analyzing a candidate version of a memory-augmented QA system.

Your task is to evaluate the candidate version against its selected base version after full training-set execution and decide how it should be used in the version tree.

\textbf{AVAILABLE TOOLS:}

\begin{itemize}[leftmargin=1.5em, itemsep=0.2em, topsep=0.2em]
    \item \texttt{\{Tool BenchmarkRunner: execute the candidate version on the training split and collect aggregate results.\}}
    \item \texttt{\{Tool ResultReader: read score summaries, success rates, token usage, and per-example outcomes.\}}
    \item \texttt{\{Tool TraceInspector: inspect weak or regressed cases through predictions, retrieved evidence, memory outputs, and execution traces.\}}
    \item \texttt{\{Tool RegistryWriter: record the version's change summary, performance analysis, and next-step recommendation.\}}
\end{itemize}

\textbf{ANALYSIS CRITERIA:}

\begin{itemize}[leftmargin=1.5em, itemsep=0.2em, topsep=0.2em]
    \item Use the primary metric as the main decision signal.
    \item Compare the candidate against its selected base version under the same training setting.
    \item Consider success rate, failure count, regression severity, and improvements on previously weak examples.
    \item Inspect weak and regressed cases only to identify broad remaining failure modes.
    \item Consider token usage only when performance is tied or nearly tied.
\end{itemize}

\textbf{DECISION AND OUTPUT:}

Accept the version as a new strong base if it improves performance with acceptable regression risk. Keep it as a side branch if it shows a useful direction but is not clearly better overall. Recommend another edit--debug round if the change is promising but incomplete. Discard it if it weakens performance, increases instability, or improves only diagnostic traces without improving the primary metric.

Write a concise analysis note covering the candidate version, parent/base relationship, what changed, whether memory was reused or rebuilt, performance comparison, major improvements and regressions, remaining failure patterns, final decision, and recommended next experiment. Update the version registry with the same conclusion.

Do not use held-out evaluation information or gold-answer shortcuts.

\end{tcolorbox}

\caption{Evaluation-stage prompt used by the \textit{Evolving Agent} to evaluate the newly expanded MCR version and generate its evaluation log.}

\label{app:eval_prompt}
\end{figure*}

\end{document}